\documentclass[10pt,twocolumn,letterpaper]{article}
\usepackage[accsupp]{axessibility}  

\usepackage{cvpr}              

\definecolor{cvprblue}{rgb}{0.21,0.49,0.74}
\usepackage[pagebackref,breaklinks,colorlinks,allcolors=cvprblue]{hyperref}

\usepackage{multirow}
\usepackage{colortbl}
\usepackage{amssymb} 

\def\paperID{11721} 
\def\confName{CVPR}
\def\confYear{2025}

\title{D$^3$: Scaling Up Deepfake Detection by Learning from Discrepancy}

\author{
    Yongqi Yang$^{1}$\thanks{Equal contribution.}, Zhihao Qian$^{1}$$^{*}$, Ye Zhu$^{2}$, Olga Russakovsky$^{2}$, Yu Wu$^{1}$\thanks{Corresponding author.}\\
    \small{\textsuperscript{1} School of Computer Science, Wuhan University, \textsuperscript{2} Department of Computer Science, Princeton University}\\
    {\tt\small \{yongqiyang, qianzhihao, wuyucs\}@whu.edu.cn, \{yezhu, olgarus\}@princeton.edu}
}

\begin{document}
\maketitle
\begin{abstract}
The boom of Generative AI brings opportunities entangled with risks and concerns.
Existing literature emphasizes the generalization capability of deepfake detection on unseen generators, significantly promoting the detector's ability to identify more universal artifacts.
This work seeks a step toward a universal deepfake detection system with better generalization and robustness. 
We do so by first scaling up the existing detection task setup from the one-generator to multiple-generators in training, during which we disclose two challenges presented in prior methodological designs and demonstrate the divergence of detectors' performance.
Specifically, we reveal that the current methods tailored for training on one specific generator either struggle to learn comprehensive artifacts from multiple generators or sacrifice their fitting ability for seen generators (i.e., \emph{In-Domain} (ID) performance) to exchange the generalization for unseen generators (i.e., \emph{Out-Of-Domain} (OOD) performance). 
To tackle the above challenges, we propose our \textbf{D}iscrepancy \textbf{D}eepfake \textbf{D}etector (\textbf{D$^\textbf{3}$}) framework, whose core idea is to deconstruct the universal artifacts from multiple generators by introducing a parallel network branch that takes a distorted image feature as an extra discrepancy signal and supplement its original counterpart. 
Extensive scaled-up experiments demonstrate the effectiveness of \textbf{D$^\textbf{3}$}, achieving 5.3\% accuracy improvement in the OOD testing compared to the current SOTA methods while maintaining the ID performance. 
The source code will be updated in our GitHub repository: \url{https://github.com/BigAandSmallq/D3}.
\end{abstract}    
\section{Introduction}
\label{sec:intro}

Recent advances in generative modeling have marked a new era of Generative AI, which is transforming society as it progresses from our research community into the real world. 
The widespread appeal of generative AI technologies among the general populace can be attributed to the significant achievements in scaling generative models—such as BigGAN~\cite{biggan}, DALL$\cdot$E~\cite{dalle}, and StableDiffusion~\cite{ldm} — to unprecedented model sizes and leveraging expansive, open-world datasets. 
However, with these opportunities come inherent risks; these advancements also raise concerns about diverse aspects including data privacy, AI ethics, social fairness, and regulatory compliance~\cite{epstein2023art}. 
Among these challenges, the risk of malicious use of synthetic data stands out as one of the particularly pressing issues that demand focused research attention.

Deepfake detection, which aims at discerning between real and generated fake images, acts as a crucial defense in responsibly deploying large generative models in the real world. 
The effectiveness of these detection systems largely depends on their generalization capability; ideally, they should reliably identify any synthesized image as fake, regardless of the generating source (i.e., the deep generative models from which they are generated).
Accordingly, current deepfake detection methods~\cite{ufd,cnndet,lnp,patchfor,dire,rasingthebar} usually follow a standard task setup for evaluating performance, which involves training on images from one specific generator and testing on images from various unseen generators, emphasizing and enhancing the detector's ability to identify universal artifacts from a single generator source.
As a natural step towards a universal deepfake detection foundation model, in this work, we propose to \textbf{scale up} this current setup from \emph{``train-on-one and test-on-many''} to \emph{``train-on-many and test-on-many''}, to accommodate the rapid development pace from the generative modeling side and better align with the real-world scenario.

However, in the process of scaling up existing deepfake detection models to multiple generators, existing methods encounter \textbf{two challenges that have not yet been adequately addressed in prior literature,} limiting the generalization performance and robustness of an expanded detection system.
\emph{First}, we reveal that the existing methodology designs tailored for the \emph{``train-on-one and test-on-many''} setup \textbf{struggle to learn the comprehensive and universal artifacts} when the fake images in training present different fingerprint patterns from more than one generator. 
For instance, one of the current SOTA detection models, DIRE from ICCV'23~\cite{dire}, tends to learn the most obvious generator-specific artifacts (e.g., fingerprints of diffusion models) but overlooks the subtle but invariant common artifacts shared by other generative AI models (e.g., fingerprints from GAN-based generators), leading to inferior testing performance on the images produced by unseen generators.  
As detailed in \cref{sec:exp}, our experimental results show that, when scaling up to 8 training generators and 12 unseen generators, DIRE~\cite{dire} achieves 97.6\% on in-domain (ID) testing, but performs significantly worse on the critical out-of-domain (OOD) testing with only 68.4\% accuracy.
\emph{Second}, our empirical analysis further suggests that \textbf{prior methods tend to comprise the trade-off between training and testing performance}, by ``underfitting'' the ID training to barter for better OOD generalization ability in testing. 
As a concrete example, another SOTA detection method, UFD from CVPR'23~\cite{ufd}, achieves better generalization on OOD testing with an accuracy of 81.4\% compared to DIRE~\cite{dire}. 
However, the better OOD performance of UFD comes with the expense of the lower ID testing performance, resulting in only 86.6\% on the ID accuracy. 

Furthermore, Detectors exhibit generalization performance divergence when scaled from one to eight generators. As illustrated in \cref{multigenerator}, Patchfor~\cite{patchfor}, initially slightly weaker than CNNDet~\cite{cnndet}, greatly surpasses CNNDet by 9\% after scaling up. This suggests that the proposed new setup intrinsically demands a comprehensive ability to acquire universal image artifacts across multiple generators for achieving higher generalization, which is different compared to the current setup.

To tackle the above challenges on the way of scaling up the deepfake detection methods to a universal system, we introduce our \textbf{D$^\textbf{3}$} framework, terminology from \textbf{D}iscrepancy \textbf{D}eepfake \textbf{D}etector, for an extended setup under the \emph{``train-on-many and test-on-many''} scenario.
\textbf{The secret recipe of \textbf{D$^\textbf{3}$}} lies within the core idea of deconstructing the universal artifacts from the manually introduced discrepancy in image features, which intuitively facilitates the universal artifacts learning among various generators and improves the testing robustness. 
Specifically, unlike the existing methods that feed the detection model only with either real or generated images, we adopt a two-branch design to provide \textbf{D$^\textbf{3}$} with an extra distorted image corresponding to its original counterpart, as illustrated in \cref{fig:method} and detailed in \cref{sec:method}.

\textbf{Intuitively}, the distorted image, created via operations like patch shuffling, flipping, and rotation, deconstructs the unique fingerprint from a specific generator into a different level of view and serves as a discrepancy signal to supplement the original image to find the generic artifact across generators.
This unique methodological design encourages the proposed \textbf{D$^\textbf{3}$} framework to capture the different artifacts distributed in the original and distorted images respectively, and the artifact correlation between them.

We perform extensive experiments by gradually scaling up six deepfake detection methods (including \cite{ufd,patchfor,lnp,cnndet,dire} and our proposed \textbf{D$^\textbf{3}$}) from conventional setup to 8 training generators and 12 testing unseen generators, on the UniversalFakeDetect (UFD)~\cite{ufd} and the GenImage~\cite{genimage} datasets.
\emph{Each detection model is re-trained till convergence to ensure a fair comparison}.
Our comprehensive experiments and analysis explicitly disclose the two previously mentioned issues by outperforming SOTA methods with 5.3\% on the OOD testing mean accuracy while maintaining the ID performance.

In summary, our work has the following contributions:
\begin{itemize}
    \item We propose to scale up the current deepfake detection setup from \emph{one-generator} to \emph{multiple-generators} in training to accommodate the real-world scenario and disclose two factors in the existing methods that limit the generalization ability and robustness in this extended detection scenario.
    \item We perform extensive experiments, by gradually scaling up 5 baselines in the current setup to 8 generators, showing the intrinsic requirements for achieving higher generalization ability in this new setup are different from the current setup.
    \item We introduce a novel framework \textbf{D}eepfake \textbf{D}etector (\textbf{D$^\textbf{3}$}) to address the above challenges, whose core idea is to learn the general artifacts among multiple generators with an extra distorted image as the discrepancy signal via a parallel network branch. Our method achieves 5.3\% improvement in the OOD testing performance compared to SOTA methods while maintaining the ID performance.
\end{itemize}

\section{Related Work}
\label{sec:related_work}
With the swift advancement of generative models~\cite{gan,kingma2013auto,ddpm,song2020score,ldm,guideddiffusion,saharia2022photorealistic} and customization methods~\cite{ruiz2023dreambooth,gal2022image,cow,zhu2024boundary,kim2022diffusionclip,controlnet}, synthesized images are becoming increasingly realistic, making it challenging for the human eye to differentiate between real and fake images. 
To counter potential malicious usage, a considerable amount of research has been devoted to detecting generated images, yielding impressive identification accuracy. 

\subsection{Artifact-based Detection Methods}
These methods explicitly extract artifacts in generated images in a preprocessing way.
They usually directly estimate the artifacts \cite{autogan,fingerprintnet}, compute the residual noise between the original image and its reconstructed one \cite{marra2019gans, lnp, dire}, or calculate gradients \cite{lgrad}.
\cite{marra2019gans} was the first to reveal that GANs leave specific fingerprints on the generated images. This was achieved by extracting the noise residual through an appropriate denoising filter, demonstrating its potential for forensic analysis.
LNP~\cite{lnp} pretrain a reconstruction network with only real images to extract the Learned Noise Patterns (LNP), which can explicitly embody artifacts.
DIRE~\cite{dire} used the prior from a pre-trained diffusion model to determine whether an image is within the distribution of fake images. This was achieved by computing the Diffusion Reconstruction Error (DIRE).

However, their artifact extraction and simulation capabilities are limited by the prior knowledge derived solely from particular models.
Therefore, they can not deal with samples generated by unfamiliar generators, making it difficult to be universally applied to real-world applications.

\subsection{Learning-based Detection Methods}
This type of method considers directly learning the artifacts from raw images. 
They aim to design a better network to extract more discriminative features.
CNNDet~\cite{cnndet} directly trained ResNet50 as the classifier on their proposed semantic-aligned dataset, containing fake images generated by ProGAN.
Their methods exhibited strong generalization capabilities to other CNN-based generative models.
Patchfor~\cite{patchfor} used a fully convolutional patch-based classifier, emphasizing local patches over global structure.
However, the learning-based methods directly train the entire network to learn the presence or absence of actual artifact patterns.
They often overlook features inherent to real images, which can lead to overfitting and poor generalization.

\subsection{Pretrained-based Detection Methods}
\label{subsec:clip-detect}

To mitigate the shortcomings of the learning-based methods, this type of method leverages the frozen feature space to extract more generalizable features.
UFD~\cite{ufd} first highlighted that a feature space not explicitly learned for generated image detection is more effective due to its unbiased decision boundaries.
Consequently, they performed detection in the pre-trained CLIP feature space using linear probing or nearest neighbor methods, demonstrating impressive generalization performance on unseen generated images.
Taking a further step in this direction, \cite{rasingthebar} trained SVMs with the features from the penultimate layer and achieved commendable performance despite a limited amount of well-organized training data.
Likewise, \cite{defake} additionally utilized the text encoder to exploit the prompt information by concatenating the encoded image embeddings and text embeddings together as features.
However, these current Pretrained-based methods heavily rely on linear separability in the feature space to distinguish between real and fake images. Therefore, when applied to a larger multi-generator dataset, the single pre-trained feature space and simple classifier struggle to effectively fit the more complex ID data, which further impacts OOD performance.
In this work, we further enhance the fitting and generalization ability of Pretrained-based detectors on multi-generator datasets.
We achieve this by supplementing a distorted image signal to deconstruct the artifacts and learn more universal and invariant artifacts, facilitating a better understanding of the universal artifacts.
\begin{figure*}[!tp]
    \centering
    \includegraphics[width=0.8\linewidth]{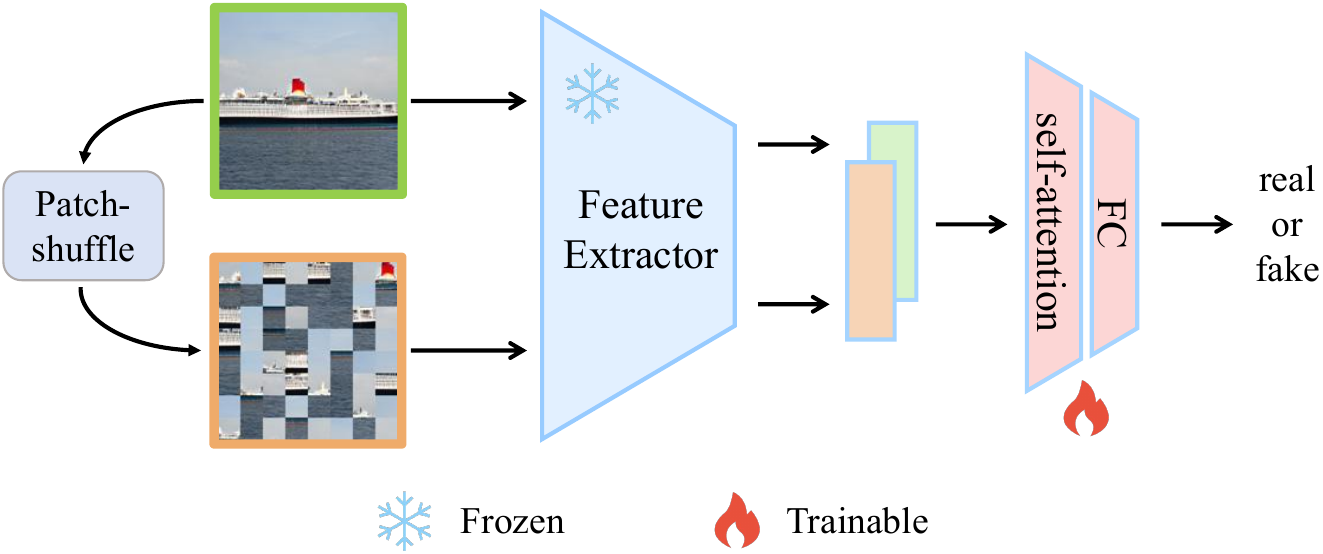}
    \caption{The overall framework of our \textbf{D}iscrepancy \textbf{D}eepfake \textbf{D}etector (\textbf{D$^\textbf{3}$}). The original image and its patch-shuffled variant are inputted into the pre-trained vision model to extract two features. We then utilize the self-attention module to encourage the learning of shared artifacts between the features. Finally, we use a single-layer linear classifier to get the final prediction.}
    \label{fig:method}
\end{figure*}

\section{Scaling Up Deepfake Detection}
\label{sec:method}

We formulate the extended \emph{``train-on-many and test-on-many''} setup in \cref{subsec:formulation}, and identify the key challenges for a generalized and robust detection system in \cref{subsec:challenges}.
\cref{subsec:framework} introduces our proposed \textbf{D$^\textbf{3}$} method.  

\subsection{Problem Formulation}
\label{subsec:formulation}

Formally, given a generator set $\mathcal{G}$ composed of $N$ different types of generators, $\mathcal{G}=\{G_i\}_1^N$, 
$\mathcal{R}_i=\{r^{(i)}_1, r^{(i)}_2, \ldots, r^{(i)}_{N_i}\}$ and $\mathcal{F}_i=\{f^{(i)}_1, f^{(i)}_2, \ldots, f^{(i)}_{N_i}\}$ denote the real and fake samples of $G_i$, where $N_i$ is the number of samples in each class.
Our goal is to correctly distinguish between real and fake when given any image from the overall dataset, $\mathcal{D}=\{\{\mathcal{R}_i\}_1^N\cup\{\mathcal{F}_i\}_1^N\}$.
The existing detection task setup involves training on one of the generators $G_i$.
In terms of evaluation, they regard the seen generator $G_i$ as the ID generator and the other unseen generators $\{G_j\}_{j \neq i}$ as the OOD generators.
This setup implies learning to detect universal artifacts from only one specific kind of artifact, which is hard and unnecessary.
This impractical setup may mislead the research community because the generalization performance is unstable and would highly depend on the choice of the training generator. For example, the OOD performances of UFD~\cite{ufd} would drop significantly from 83.2\% to 58.9\% when switching the training from the ADM~\cite{guideddiffusion} to Midjourney~\cite{midjourney} in this scheme. 

Therefore, we propose to step out of the “train-on-one and test-on-many” task setup and leverage a new \emph{“train-on-many and test-on-many”} evaluation scheme.
Specifically, we add more generators into the current dataset and train on a subset composed of $k$ $(k<N)$ generators, denoted by $\mathcal{G}_s=\{G_{s_i}\}_1^k ( s_i \in \{1,2,\ldots,N\})$.
In this way, we regard the $k$ generators in $\mathcal{G}_s$ as the ID generators and the other $N-k$ generators as the OOD generators.
In practice, we merge two large-scale existing deepfake detection datasets, i.e., the UFD dataset~\cite{ufd} and GenImage dataset~\cite{genimage}. Finally, there are in total 20 state-of-the-art generators available in the scale-up setup. 
Considering the diversity, we regard all the 8 generators used in different previous works, including 2 GANs methods like ProGAN~\cite{progan}, and 6 DMs like ADM~\cite{guideddiffusion}. 
The rest 12 generators (including both GANs and DMs) are used as the OOD set.

\subsection{Challenges in the Scaled-up Setup}
\label{subsec:challenges}
\begin{figure*}[!tp]
    \centering
    \includegraphics[width=0.7\textwidth]{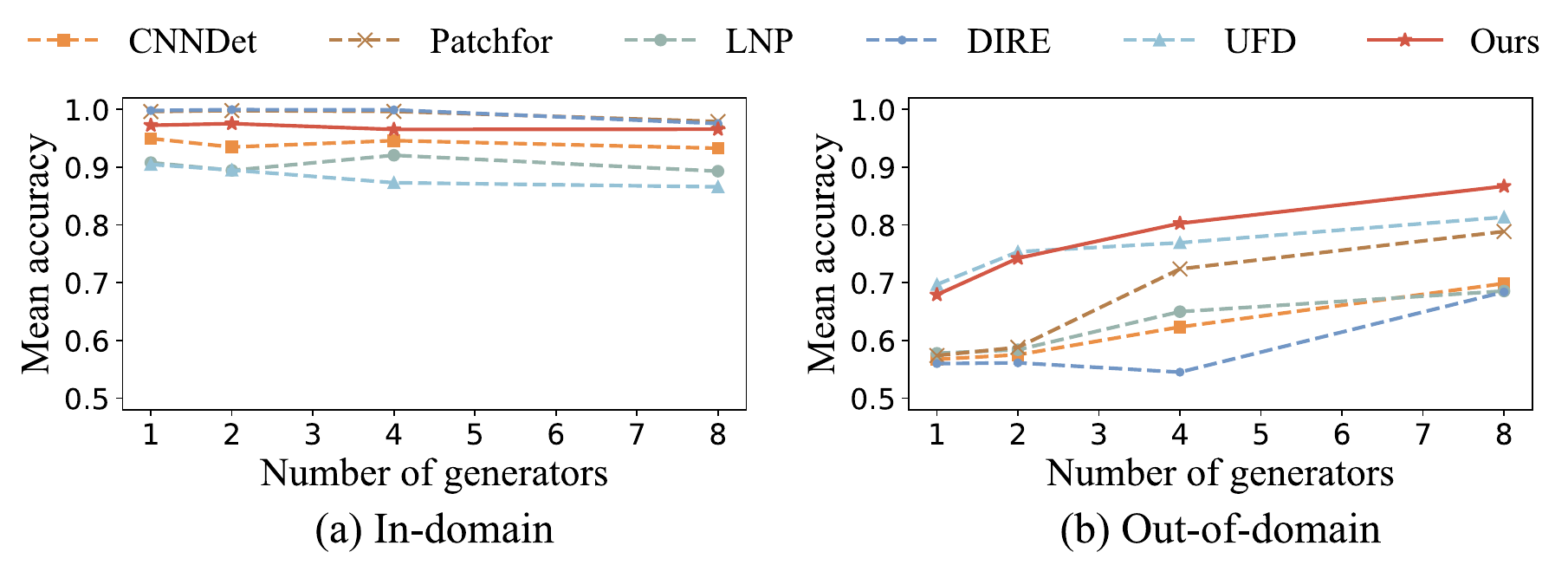}
    \caption{Mean accuracy of training on scaled-up datasets on the testing set. We design an experiment to show the performance of different methods when gradually adding generators into the training pool. To avoid occasionality, we add the generator in three distinct random orders and average the corresponding test results. (a) ``In-domain'' means 1 shared training generator (\textit{the first generator in any given order}). (b) ``Out-of-domain'' means 12 shared OOD generators.}
    \label{multigenerator}
\end{figure*}

Based on the scaled-up setup, we gradually add the generators into the training pool in different random orders to evaluate the existing works.
As the number of generators increased from one to eight, we found existing methodology designs tailored for the previous setup could not learn universal artifacts. 
For example, one of the state-of-the-art works UFD~\cite{ufd} from CVPR'23 even decreases significantly in terms of ID testing accuracy as more generators are used for training (\cref{multigenerator}), demonstrating the underfitting issue. 
While another state-of-the-art work DIRE~\cite{dire} from ICCV'23 shows extremely good ID performance (97.6\%), it suffers from the generalization issue that the OOD testing accuracy is not promising (68.4\%), despite it has already seen eight advanced generators covering typical GANs and DMs. 
These results suggest existing models could not find the invariant artifacts that are universal in multiple generators. 
The main reason might be these detectors struggle to capture the universal multi-generator artifact patterns we wish the model to learn.

In addition, detectors exhibit performance divergence when scaled from one to eight generators. For example, Patchfor, initially weaker than CNNDet and just comparable to LNP and DIRE, achieves a substantial increase to 0.789 when using eight generators. This performance greatly surpasses CNNDet (0.699), LNP (0.719), and DIRE (0.684). It indicates that the intrinsic requirements for achieving higher generalization under multi-generator training are different from the one-generator training. The scaled-up setup requires a more comprehensive ability to acquire universal image artifacts across multiple generators. This inspires us to go deeper to deconstruct the artifacts of each specific generator into different levels of view, therefore learning the universal artifacts rather than the superficial and generator-specific artifacts that can be directly seen on the deepfake images.

\subsection{\textbf{D$^\textbf{3}$}: Discrepancy-based Deepfake Detection}
\label{subsec:framework}

\textbf{Overall Pipeline.} Our proposed framework for generative image detection is depicted in \cref{fig:method}. Overall, the original image and its patch-shuffled counterpart are input into the pre-trained CLIP~\cite{clip} model to obtain visual features. 
A self-attention layer is then used to promote learning from the discrepancy between the two features. Finally, a linear classifier is employed to predict real or fake labels. The design of each module will be discussed in the following.

\noindent
\textbf{Visual Backbone.} Directly learning on the deepfake detection data without pretraining can lead to overfitting on the training generators, resulting in poor generalization to OOD generators~\cite{corvi2023detection, genimage}. To mitigate this, we follow previous work \cite{ufd,rasingthebar} and adopt the pre-trained CLIP:ViT-L/14 as our feature extractor. More specifically, we extract features from the penultimate layer of the pre-trained CLIP model, which keeps more detailed visual clues of the input.

\noindent
\textbf{Dual-Path Discrepancy Learning.}
To encourage the model to find deeper robust artifacts, we propose to introduce an additional branch that takes a corrupted image as input.
Given an original image $x_o$ from the dataset $\mathcal{D}_s=\{\{\mathcal{R}_i\}_1^k\cup\{\mathcal{F}_i\}_1^k\}$, corresponding to the subset $\mathcal{G}_s$.
In the new branch, we divide the input image $x_o$ into patches, and then randomly shuffle them to form a new image, where the superficial artifacts will be destroyed to avoid the model simply learning a generator-specific shortcut. 
After that, we input the corrupted data to the same visual backbone to extract features, 
\begin{equation}
    e_o = \texttt{CLIP}^*(\texttt{AUG}(x_o)), e_s = \texttt{CLIP}^*(\texttt{PS}(\texttt{AUG}(x_o))),
\end{equation}
where $e_o$ and $e_s$ are the visual embeddings of the respective images, $\texttt{CLIP}^*$ denotes the extraction of the penultimate feature using the pre-trained CLIP model, $\texttt{PS}$ is the patch-shuffling operation, and $\texttt{AUG}$ represents the common data augmentation method used in the previous work \cite{cnndet}. 


\noindent
\textbf{Self-Attention Invariance Extraction.} Having obtained two features, $e_o$ and $e_s$, from the original and patch-shuffled images respectively, we stack the two features together, $e = [e_o, e_s]$.  The correlation between $e_o$ and $e_s$ is crucial for detecting fake images. To this end, we employ a self-attention layer on the stacked feature $e$ to extract the shared artifact feature $E$ by learning the associations between the features. In this way, we force the model to learn from the discrepancy between the original image and its patch-shuffled image.
In experiments, we find that the more distinct the original image and the corrupted image are, the better the deepfake detection performance of the model is.

\noindent \textbf{Loss Function.} Based on the fused feature $E$, we employ a single fully connected layer, denoted as $\Phi(\cdot)$, to compute the classification logits.
We then train the model using the binary cross entropy loss by,
\begin{equation}
    \mathcal{L} = -\frac{1}{N}\sum_{i=1}^{N} (y_i\log(\Phi(E)) + (1-y_i)\log(1-\Phi(E))),
\end{equation}
where $N$ is the batch size, and $y_i$ is the label of the $i$-th input sample. 
Our backbone is frozen, thus the learnable parameters are only self-attention and the classification layer.

\section{Experiments}
\label{sec:exp}

\subsection{Scaled-Up Datasets}
\label{sec:dataset}
To evaluate the effectiveness of our proposed method, we conducted experiments on the UniversalFakeDetect dataset \cite{ufd} and the GenImage dataset \cite{genimage}. 
Notably, we scale up the experimental data by merging the two datasets together.

The \textbf{UniversalFakeDetect} (abbreviated to ``\textbf{UFD}'') dataset contains 720k images in its training set, including 360k real images (drew from LSUN~\cite{yu2015lsun}, LAION~\cite{schuhmann2021laion} and ImageNet~\cite{deng2009imagenet}) and 360k fake images generated by different generators including GANs like ProGAN~\cite{progan}, CycleGAN~\cite{cyclegan}, BigGAN~\cite{biggan}, StyleGAN~\cite{stylegan}, GauGAN~\cite{gaugan}, StarGAN~\cite{stargan}, Deepfakes~\cite{deepfakes}, SITD~\cite{SITD}, SAN~\cite{SAN}, CRN~\cite{crn} and IMLE~\cite{IMLE}, and diffusion models like Guided Diffusion~\cite{guideddiffusion}, LDM~\cite{ldm}, GLIDE~\cite{glide} and DALL$\cdot$E~\cite{dalle}.

The \textbf{GenImage} dataset comprises 2,681,167 images, segregated into 1,331,167 real and 1,350,000 fake images, with 50k images left for testing in each generator type. Fake images in this dataset are generated by 8 generators, including 1 GAN (BigGAN~\cite{biggan}) and 7 diffusion models (Stable Diffusion V1.4~\cite{ldm}, Stable Diffusion V1.5, GLIDE~\cite{glide}, VQDM~\cite{vqdm}, Wukong~\cite{wukong}, ADM~\cite{guideddiffusion} and Midjourney~\cite{midjourney}).

In our new setup, we merge two datasets into one and treat generators with the same architecture but different parameters (\textit{e.g.} SD v1.4 and SD v1.5) as the same type. 
To this end, we eliminate SD v1.5 from the merged training set and remain SD v1.4.
Besides, we make sure each generator has an equal number of samples to avoid imbalanced data. 
Finally, there are 20 generators available in the scale-up setup, 8 in the training set (2 GANs and 6 Diffusions) and 12 in the OOD set. 
We further split 10\% samples per generator of the testing set as the validation set.

\subsection{Implementation Details}
In our experiments, we employ the CLIP: VIT-L/14 model pre-trained on a large dataset WebImageText as the backbone to extract patch tokens. We adopt its penultimate feature before dimensionality reduction for better representation. 
With the backbone frozen, the whole network is trained on a single NVIDIA RTX 4090 with a batch size of 128 within 5 epochs. We apply the Adam optimizer with an initial learning rate of 0.0001 and a weight decay of 0. During training, all images are resized to 256x256 resolution and then randomly cropped to 224x224 resolution. We adopted the data augmentation scheme the same as the works~\cite{cnndet, ufd}, involving Gaussian blur and JPEG compression. Specifically, the \textit{quality} of JPEG compression randomly ranges from 30 to 100, the \textit{sigma} of Gaussian blur randomly ranges from 0 to 3, and their occurrence probabilities are set to 0.5. During validation and testing, we resize the image to size 224x224, without any augmentation.

\begin{table}[!tp]
    \centering
    \setlength{\belowcaptionskip}{3pt}
    \tabcolsep=1pt
    \resizebox{\linewidth}{!}{
    \begin{tabular}{cc|cccccc}
    \toprule
        \multirow{3}{*}{Methods} & \multirow{3}{*}{Pub.} & \multicolumn{2}{c}{In-domain} & \multicolumn{2}{c}{Out-of-domain} & \multicolumn{2}{c}{Total} \\
        ~ & ~ & Mean acc. & AP & Mean acc. & AP & Mean acc. & AP  \\
    \hline
        CNNDet      & CVPR20 & 0.933 & 0.906 & \cellcolor{gray!30}0.699 & \cellcolor{gray!30}0.679 & 0.792 & 0.770   \\ 
        Patchfor    & ECCV20 & \textbf{0.979} & \textbf{0.997} & \cellcolor{gray!30}0.789 & \cellcolor{gray!30}0.880 & 0.865 & 0.942  \\ 
        LNP         & ECCV22 & 0.881 & 0.902 & \cellcolor{gray!30}0.719 & \cellcolor{gray!30}0.797 & 0.784 & 0.823 \\
        DIRE        & ICCV23 & 0.976 & 0.994 & \cellcolor{gray!30}0.684 & \cellcolor{gray!30}0.726 & 0.801 & 0.862 \\
        UFD         & CVPR23 & 0.866 & 0.950 & \cellcolor{gray!30}0.814 & \cellcolor{gray!30}0.927 & 0.835 & 0.935  \\
    \hline
        Ours        & -      & 0.966 & 0.995 & \cellcolor{gray!30}\textbf{0.867} & \cellcolor{gray!30}\textbf{0.943} & \textbf{0.907} & \textbf{0.968}  \\
    \bottomrule
    \end{tabular}}
    \caption{Comparisons to the SOTAs in mean accuracy (Mean acc.) and average precision (AP) on the testing set. ``In-domain'' includes 8 types of generators seen during training, ``Out-of-domain'' includes 12 types of generators, and ``Total'' includes all of them. Bold represents the best within the same metric. Our method achieves the best OOD and overall performance. \textbf{Note that UFD uses the same backbone as ours.} A more detailed version of the result can be found in the appendix.}
    \label{tab:sota}
\end{table}

\subsection{Experimental Setup}
\label{exp:setup}
In this subsection, we provide a detailed illustration of our experimental setup, including the implementation of baselines and evaluation metrics.

\noindent
\textbf{Implementation of Baselines.}
In terms of baselines, we \textbf{re-train} the comparison methods according to their official codes under our problem settings for fairness. The baselines include CNNDet~\cite{cnndet} (ResNet50 version), Patchfor~\cite{patchfor} (Xception version), LNP~\cite{lnp}, DIRE~\cite{dire}, and UFD~\cite{ufd} (linear probing version). In our experiments, all methods are re-implemented with their default settings in our new setup, and the optimal checkpoint is selected based on the validation set performance to ensure a fair comparison.

\noindent
\textbf{Evaluation Metrics.}
We employ mean classification accuracy (Mean acc.) and average precision (AP) to evaluate the performance of detectors. When calculating overall mean accuracy, we first average the accuracy of different variants of the same generator (\textit{e.g.} SD1.4 and SD1.5). Then, we average all the accuracy scores across different generators. Note that we use a fixed threshold of 0.5 when calculating the accuracy score for each detector. 

AP quantifies the separability between fake and real images, neglecting the specific threshold consideration. 
Previous literature~\cite{ufd, lnp, dire,cnndet} typically uses mAP to represent the overall AP score, where they calculate AP for each generator independently and average the AP scores.
However, the direct average of AP disregards the substantial variance in the optimal thresholds, or boundaries, between real and fake images from different generators. 
In practice, it is infeasible for users to set different real/fake thresholds for unseen generators. Therefore, we propose to calculate a global AP rather than mAP, which shares the same standards among all testing generators.

We first resample generators with the same architecture based on the maximum number of samples among them and merge them as one.
Then, we similarly resample different types of generators to ensure the entire test set is evenly distributed.
Finally, we calculate the average precision (AP) of different merged sets to represent the overall AP. 
For instance, in terms of ID AP, we merge the results of all ID generators to compute an overall ID AP.
This approach ensures a balanced evaluation across different generators, providing a more accurate measure of the detector's performance.

\subsection{Scaling Up the Training Data}\label{sec:multigen}
In this section, we conduct experiments to show how the models perform as the generators gradually scale up in \cref{multigenerator}.
We re-train the detectors on the training set with \textit{\{1, 2, 4, 8\}} generators. To mitigate any potential bias, we add the generator in three distinct random orders and average the corresponding test results.  To make all the curves comparable (in both horizontal and vertical), all detectors are evaluated on the same ID and OOD test sets, thus the curves are comparable in both horizontal and vertical and we can observe how a model behaves as the number of training generators grows.
Note that only the first generator is commonly shared when gradually adding training generators from one to eight, we thus test all models only on this generator for the ID evaluation.
For the OOD evaluation, we use the 12 common unseen generators for all models.

According to \cref{multigenerator}, as the number of generators increases, our method shows a slight fluctuation in ID performance and exhibits the highest performance and stable upward trend in OOD generalization.
CNNDet~\cite{cnndet} and Patchfor~\cite{patchfor} show excellent ID fitting ability, but their OOD performances are inferior to the other works. 
It is worth mentioning that Patchfor to some extent alleviates the overfitting issue by focusing on artifacts of local patches instead of the high-level features.
Reconstruction-based methods like LNP and DIRE heavily rely on their pre-trained reconstruction model, which could result in poor generalization to unfamiliar domains.
Moreover, the pretrained-based method UFD exhibits superior OOD performance when there are few generators and maintains a rising trend as the training set scales up. However, its ID performance continues to decline, which indicates a limitation in its representative capacity.
These findings demonstrate that $D^3$ effectively improves the distinguishability and generalizability of features by patch-shuffling and artifact invariance learning from the introduced discrepancy, resulting in outstanding performance of ID fitting and OOD generalization.

\begin{figure*}[!tp]
    \centering
    \includegraphics[width=0.7\linewidth]{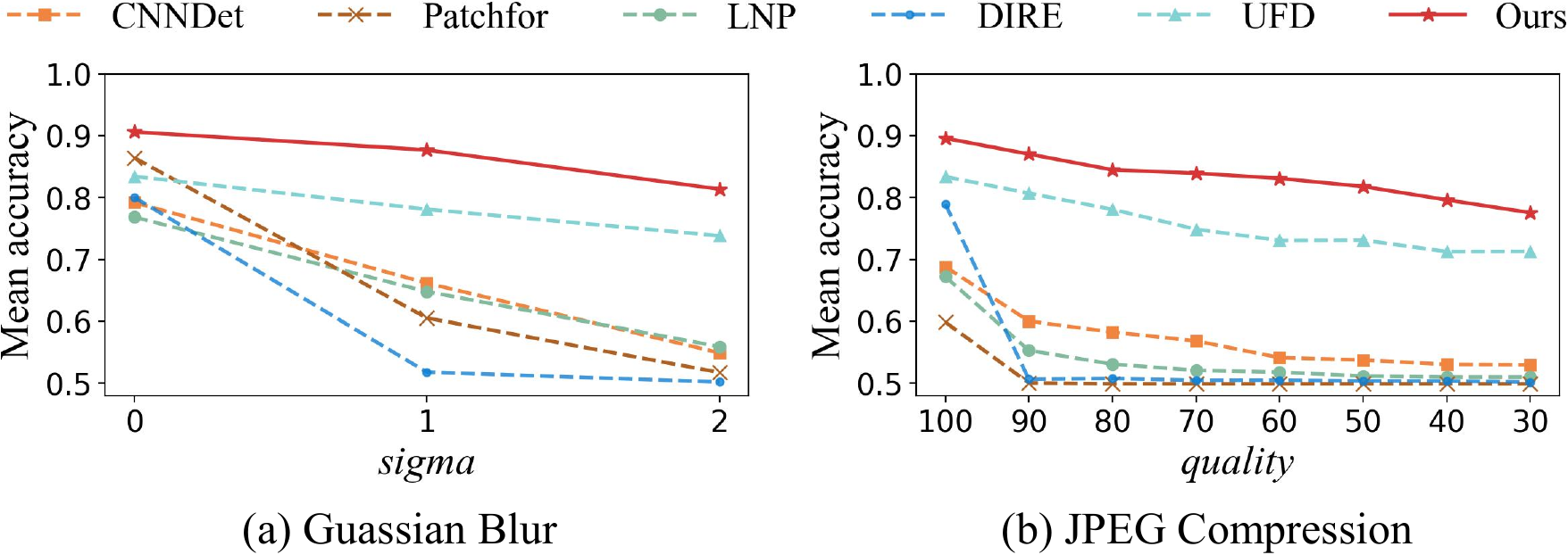}
    \caption{Total mean accuracy results of robustness to post-processing operations on the testing set. We conducted experiments on Gaussian blurring ranging from 0 to 2 and JPEG compression ranging from 30 to 100 to verify the robustness of different methods.}
    \label{robustness}
\end{figure*}

\subsection{Comparison with the State-of-the-Art Methods}
\label{sec:comparison}
We compare our method with other state-of-the-art methods in the scale-up setup and report an evaluation of their ID and OOD performance on the testing set. 
\cref{tab:sota} reports the results on the testing set, where 
``In-domain'' includes 8 types of generators, and ``Out-of-domain'' includes 12 types of generators, and ``Total'' includes all of them.
A more detailed version of the result can be found in the appendix.

\textbf{The experimental results exhibit the strengths and weaknesses of each method after scaled-up.}
CNNDet overfits data from familiar generators and performs poorly on data from unseen generators, consistent with the findings about ResNet in the paper \cite{genimage}. 
Patchfor achieves the best ID mean accuracy and AP. However, its subpar OOD performance suggests that directly learning the specific local artifact patterns may also lead to a higher potential for overfitting. LNP exhibits poor performance in both ID and OOD fitting.
DIRE demonstrates superior ID performance, surpassed only by Patchfor. However, it lacks generalization capabilities with OOD generators due to its tailored design for DMs.
UFD ranks second only to our method. However, its ID performance is significantly inferior.

\textbf{D$^\textbf{3}$ achieve the best overall performance under a fair comparison.}
In comparison, our method achieves the highest overall mean accuracy of 90.7\% and AP of 96.8\%, exceeding the second-best method, Patchfor, in overall mean accuracy by 4.2\% and AP by 2.6\%. 
Although our method marginally trails Patchfor in terms of ID accuracy and AP, it exhibits a more balanced and consistent overall performance.
Furthermore, our method greatly outperforms UFD, \textbf{which also leverages the same pre-training vision backbone}, in ID mean accuracy by 10\% and OOD mean accuracy by 5.3\%.
This underscores that our method further exploits the potential of pretrained-based feature representation ability by learning from the discrepancy of the shared artifact in the original image and its patch-shuffled variant.

\noindent
\textbf{Robustness Evaluation.} We evaluate the robustness of all methods to align more closely with their practical effectiveness in real-world applications.
Given the potential for real-world images to undergo various degradations, ensuring the robustness of detectors against unseen perturbations becomes another vital aspect of evaluating their performance. 
A detector that can resist various perturbations and attacks has higher potential and value for practical application.
Therefore, we evaluate all methods on two post-processing operations, Gaussian blurring (\textit{sigma}: 0-2) and JPEG compression (\textit{quality}: 30-100). 
\cref{robustness} presents the results of our comparison with the other methods, demonstrating that our method exhibits the best robustness with a large margin with the other methods.
Note that except for Patchfor and DIRE, all the other methods employ an augmentation strategy involving random Gaussian blurring and JPEG compression during training to enhance robustness.
Despite this, we can observe that only our method and UFD exhibit general robustness, even to extremely heavy degradations.
In addition, from the perspective of accuracy and the decay rate, our method shows obvious advantages compared to UFD.
Discrepancy learning between the original image and its patch-shuffled one offsets the impact of these disturbances due to its concentration on more stable and invariant artifacts instead of specific artifact patterns.

\begin{table}[!tp]
    \centering
    \setlength{\belowcaptionskip}{3pt}
    \tabcolsep=1pt
    \resizebox{1\linewidth}{!}{
    \begin{tabular}{c|ccccc|cccccc}
    \toprule
        \multirow{2}{*}{Group} & \multirow{2}{*}{Branch1} & \multirow{2}{*}{Branch2} & \multirow{2}{*}{CLIP} & \multirow{2}{*}{SA} & \multirow{2}{*}{FC} & \multicolumn{2}{|c}{In-domain} & \multicolumn{2}{c}{Out-of-domain} & \multicolumn{2}{c}{Average} \\ 
        & & & & & & Mean acc. & AP & Mean acc. & AP & Mean acc. & AP\\
    \hline
        1 & Original  & -           & \checkmark    & ~             & \checkmark & 0.889 & 0.965 & 0.823 & 0.937 & 0.850 & 0.947 \\
        2 & Original  & -           & \checkmark    & \checkmark    & \checkmark & 0.894 & 0.965 & 0.834 & 0.936 & 0.858 & 0.947 \\
        3 & Original  & Shuffled    & \checkmark    & ~             & \checkmark & 0.918 & 0.977 & 0.835 & \textbf{0.938} & 0.868 & 0.954 \\
        4 & Original  & Shuffled    & \checkmark    & \checkmark    & \checkmark & \textbf{0.972} & \textbf{0.995} & \textbf{0.874} & 0.931 & \textbf{0.913} & \textbf{0.962} \\
    \hline
        5 & Original  & Original    & \checkmark    & \checkmark    & \checkmark & 0.889 & 0.964 & 0.829 & 0.935 & 0.853 & 0.946 \\
        6 & Shuffled  & Shuffled    & \checkmark    & \checkmark    & \checkmark & 0.841 & 0.914 & 0.738 & 0.800 & 0.780 & 0.849 \\
    \bottomrule
    \end{tabular}}
    \caption{Results of ablation study on the validation set. We break down our framework into several key modules and explore the importance of each module through ablation experiments. The checkmark represents the adoption of the corresponding module. The emptiness of Branch 2 means using only one feature.}
    \label{ablation}
\end{table}

\subsection{Exploration of Different Disruptions}
In this subsection, we investigate how different levels of disruptions influence model performance.  
We conduct experiments by only replacing the patch-shuffling operation with other disruption operations in our framework.
These operations include horizontal flipping, vertical flipping, and random rotating (randomly rotating an image from 0° to 180°).

The experimental results shown in\cref{augmentations}, show that each disrupted feature helps in promoting the model performance, with the effectiveness of patch-shuffling, random rotating, vertical flipping, and horizontal flipping diminishing in sequence. 
The two-branch model with the \textbf{horizontal flipping} doesn't show superior to the one-branch baseline, because the horizontally-flipped image could not provide any discrepancy and its features are just the same as the original image for most existing vision models.
Comparatively, \textbf{vertical flipping} has provided corruptions to the vision backbones to some degree since most vision models could not well recognize the image after vertical flipping. Thus the discrepancy helps the model learn better artifacts, leading to an overall performance boost of 4\% on mean accuracy compared to the horizontal flipping.
In addition, \textbf{random rotation} randomly changes the angle of images, introducing a slightly stronger semantic corruption on image semantics than the fixed 180° angle of vertical flipping, and thus has a slightly better overall performance.
In comparison to the three operations, the \textbf{patch-shuffling} operation disrupts the high-level image semantics more completely while preserving low-level information, leading to an informative discrepancy between the two branches.
Thus it achieves the best performance.
The comparison between different transformations illustrates that the discrepancy between the original images and corrupted images helps the detector learn more universal artifacts. The stronger the corruption the operation causes, the better the model performs.

\begin{table}[!tp]
    \centering
    \setlength{\belowcaptionskip}{3pt}
    \tabcolsep=0.5pt
    \resizebox{\linewidth}{!}{
    \begin{tabular}{c|cccccc}
    \toprule

        \multirow{3}{*}{Augmentations} & \multicolumn{2}{|c}{In-domain} & \multicolumn{2}{c}{Out-of-domain} & \multicolumn{2}{c}{Total} \\  
        ~ & Mean acc. & AP & Mean acc. & AP & Mean acc. & AP\\
    \hline
        Horizontal flipping & 0.887 & 0.964 & 0.819 & 0.934 & 0.846 & 0.945   \\
        Vertical flipping   & 0.934 & 0.986 & 0.854 & \textbf{0.937} & 0.886 & 0.959   \\
        Random rotation     & 0.954 & 0.991 & 0.854 & 0.928 & 0.891 & 0.958   \\
        Patch shuffling     & \textbf{0.972} & \textbf{0.995} & \textbf{0.874} & 0.931 & \textbf{0.913} & \textbf{0.962}   \\
    \bottomrule
    \end{tabular}}
    \caption{Results of different disruptions on the validation set to show how different transformations influence model performance.}
    \label{augmentations}
\end{table}

\subsection{Ablation Studies}

In this subsection, we break down our framework into several key modules and explore the role of each module through ablation experiments. 
The ablation experimental results are shown in \cref{ablation}, where a tick represents the adoption of this module and the emptiness of Branch2 means using only one branch. 
The experiments have been numbered with groups for the following reference.

\noindent
\textbf{Patch-shuffling Operation.} 
A comparison between Group 2 and Group 4 reveals that the integration of the original image and the patch-shuffled image results in a substantial performance enhancement under the ``SA''+``FC'' classifier. 
This is evidenced by an increase of 7.8\% in ID mean accuracy, 4\% in OOD mean accuracy, and 5.5\% in total mean accuracy. 
This effectively substantiates that combining the information from the original image and its patch-shuffled one can enhance the overall performance.

\noindent
\textbf{Discrepancy between Branches.}  
Compared to Group 4, both Group 5 and Group 6 show an inferior performance, due to the absence of discrepancy features between two images of the same type.
According to the table, Group 5 and Group 6 are 6.0\% and 13.3\% lower than group 4 in average mean accuracy, respectively.
This further demonstrates that different inputs can provide different information, and assist the model to learn more universal artifacts.

\noindent
\textbf{Self-Attention Layer.} 
Comparing Group 1 with Group 2, adding a self-attention layer to the single original feature only slightly improves the performance by approximately 1 point.
However, comparing Group 3 with Group 4, adding a self-attention layer to two different features markedly enhances the performance with an increase of 5.5\% in ID mean accuracy and 3.9\% in OOD mean accuracy.
This suggests that appropriately learning the connections between the shared artifacts in the original image and its patch-shuffled one, rather than merely enhancing the features, is another key to the success of our method.

In conclusion, each module is dispensable. The impact of any module may not be substantial, but their collective integration significantly enhances the performance. These findings provide robust evidence to support our proposition that learning from discrepancy is beneficial for universal deepfake detection in a scaled-up scenario.

\section{Conclusion and Discussion}
\textbf{Conclusion.} In this paper, we propose a novel evaluation setup \emph{``train-on-many and test-on-many''}, aimed at advancing the development of a universal deepfake detection system.
To tackle the challenges in the new setup, we introduce our \textbf{D}iscrepancy \textbf{D}eepfake \textbf{D}etector (\textbf{D$^\textbf{3}$}) to learn the universal artifacts from multiple generators. We achieve this by introducing a parallel network branch that takes a distorted image as an extra discrepancy signal to supplement its original counterpart. Extensive experiments demonstrate the effectiveness of \textbf{D$^\textbf{3}$}, achieving 5.3\% accuracy improvement in the OOD testing compared to the current SOTA methods and maintaining excellent ID performance.

\noindent
\textbf{Limitations.} The parallel design of \textbf{D$^\textbf{3}$} requires running the vision model twice to obtain two different features from one image, which doubles the computation cost compared with UFD. 
Our proposed setup demands a large-scale and comprehensive dataset, while our merged dataset may not fully meet this criterion in the future. Collecting such a high-quality detection dataset may be expensive, however, not infeasible given the fast-increasing research scale.

\newpage
\section{Acknowledgements}
This work was supported by the National Natural Science Foundation of China under grant 62372341. 
Research reported in this was partially supported by an Amazon Research Award Spring 2024 to OR and YZ. Any opinions, findings, and conclusions or recommendations expressed in this material are those of the authors and do not necessarily reflect the views of Amazon.

{
    \small
    \bibliographystyle{ieeenat_fullname}
    \bibliography{main}

\begin{thebibliography}{51}
\providecommand{\natexlab}[1]{#1}
\providecommand{\url}[1]{\texttt{#1}}
\expandafter\ifx\csname urlstyle\endcsname\relax
  \providecommand{\doi}[1]{doi: #1}\else
  \providecommand{\doi}{doi: \begingroup \urlstyle{rm}\Url}\fi

\bibitem[whi(2019)]{whichface}
http://www.whichfaceisreal.com/, 2019.

\bibitem[mid(2022)]{midjourney}
https://www.midjourney.com/home/, 2022.

\bibitem[wuk(2022)]{wukong}
https://xihe.mindspore.cn/modelzoo/wukong, 2022.

\bibitem[Brock et~al.(2019)Brock, Donahue, and Simonyan]{biggan}
Andrew Brock, Jeff Donahue, and Karen Simonyan.
\newblock Large scale gan training for high fidelity natural image synthesis.
\newblock \emph{ICLR}, 2019.

\bibitem[Chai et~al.(2020)Chai, Bau, Lim, and Isola]{patchfor}
Lucy Chai, David Bau, Ser-Nam Lim, and Phillip Isola.
\newblock What makes fake images detectable? understanding properties that generalize.
\newblock In \emph{ECCV}, 2020.

\bibitem[Chen et~al.(2018)Chen, Chen, Xu, and Koltun]{SITD}
Chen Chen, Qifeng Chen, Jia Xu, and Vladlen Koltun.
\newblock Learning to see in the dark.
\newblock In \emph{CVPR}, 2018.

\bibitem[Chen and Koltun(2017)]{crn}
Qifeng Chen and Vladlen Koltun.
\newblock Photographic image synthesis with cascaded refinement networks.
\newblock In \emph{ICCV}, 2017.

\bibitem[Choi et~al.(2018)Choi, Choi, Kim, Ha, Kim, and Choo]{stargan}
Yunjey Choi, Minje Choi, Munyoung Kim, Jung-Woo Ha, Sunghun Kim, and Jaegul Choo.
\newblock Stargan: Unified generative adversarial networks for multi-domain image-to-image translation.
\newblock In \emph{CVPR}, 2018.

\bibitem[Corvi et~al.(2023)Corvi, Cozzolino, Zingarini, Poggi, Nagano, and Verdoliva]{corvi2023detection}
Riccardo Corvi, Davide Cozzolino, Giada Zingarini, Giovanni Poggi, Koki Nagano, and Luisa Verdoliva.
\newblock On the detection of synthetic images generated by diffusion models.
\newblock In \emph{ICASSP}, 2023.

\bibitem[Cozzolino et~al.(2023)Cozzolino, Poggi, Corvi, Nie{\ss}ner, and Verdoliva]{rasingthebar}
Davide Cozzolino, Giovanni Poggi, Riccardo Corvi, Matthias Nie{\ss}ner, and Luisa Verdoliva.
\newblock Raising the bar of ai-generated image detection with clip.
\newblock \emph{arXiv}, 2023.

\bibitem[Dai et~al.(2019)Dai, Cai, Zhang, Xia, and Zhang]{SAN}
Tao Dai, Jianrui Cai, Yongbing Zhang, Shu-Tao Xia, and Lei Zhang.
\newblock Second-order attention network for single image super-resolution.
\newblock In \emph{CVPR}, 2019.

\bibitem[Deng et~al.(2009)Deng, Dong, Socher, Li, Li, and Fei-Fei]{deng2009imagenet}
Jia Deng, Wei Dong, Richard Socher, Li-Jia Li, Kai Li, and Li Fei-Fei.
\newblock Imagenet: A large-scale hierarchical image database.
\newblock In \emph{CVPR}, 2009.

\bibitem[Dhariwal and Nichol(2021)]{guideddiffusion}
Prafulla Dhariwal and Alexander Nichol.
\newblock Diffusion models beat gans on image synthesis.
\newblock \emph{NeurIPS}, 2021.

\bibitem[Dosovitskiy et~al.(2021)Dosovitskiy, Beyer, Kolesnikov, Weissenborn, Zhai, Unterthiner, Dehghani, Minderer, Heigold, Gelly, et~al.]{vit}
Alexey Dosovitskiy, Lucas Beyer, Alexander Kolesnikov, Dirk Weissenborn, Xiaohua Zhai, Thomas Unterthiner, Mostafa Dehghani, Matthias Minderer, Georg Heigold, Sylvain Gelly, et~al.
\newblock An image is worth 16x16 words: Transformers for image recognition at scale.
\newblock \emph{ICLR}, 2021.

\bibitem[Epstein et~al.(2023)Epstein, Hertzmann, of~Human~Creativity, Akten, Farid, Fjeld, Frank, Groh, Herman, Leach, et~al.]{epstein2023art}
Ziv Epstein, Aaron Hertzmann, Investigators of Human~Creativity, Memo Akten, Hany Farid, Jessica Fjeld, Morgan~R Frank, Matthew Groh, Laura Herman, Neil Leach, et~al.
\newblock Art and the science of generative ai.
\newblock \emph{Science}, 2023.

\bibitem[Gal et~al.(2023)Gal, Alaluf, Atzmon, Patashnik, Bermano, Chechik, and Cohen-Or]{gal2022image}
Rinon Gal, Yuval Alaluf, Yuval Atzmon, Or Patashnik, Amit~H Bermano, Gal Chechik, and Daniel Cohen-Or.
\newblock An image is worth one word: Personalizing text-to-image generation using textual inversion.
\newblock \emph{ICLR}, 2023.

\bibitem[Goodfellow et~al.(2014)Goodfellow, Pouget-Abadie, Mirza, Xu, Warde-Farley, Ozair, Courville, and Bengio]{gan}
Ian Goodfellow, Jean Pouget-Abadie, Mehdi Mirza, Bing Xu, David Warde-Farley, Sherjil Ozair, Aaron Courville, and Yoshua Bengio.
\newblock Generative adversarial nets.
\newblock \emph{NeurIPS}, 2014.

\bibitem[Gu et~al.(2022)Gu, Chen, Bao, Wen, Zhang, Chen, Yuan, and Guo]{vqdm}
Shuyang Gu, Dong Chen, Jianmin Bao, Fang Wen, Bo Zhang, Dongdong Chen, Lu Yuan, and Baining Guo.
\newblock Vector quantized diffusion model for text-to-image synthesis.
\newblock In \emph{CVPR}, 2022.

\bibitem[Ho et~al.(2020)Ho, Jain, and Abbeel]{ddpm}
Jonathan Ho, Ajay Jain, and Pieter Abbeel.
\newblock Denoising diffusion probabilistic models.
\newblock \emph{NeurIPS}, 2020.

\bibitem[Jeong et~al.(2022)Jeong, Kim, Ro, Kim, and Choi]{fingerprintnet}
Yonghyun Jeong, Doyeon Kim, Youngmin Ro, Pyounggeon Kim, and Jongwon Choi.
\newblock Fingerprintnet: Synthesized fingerprints for generated image detection.
\newblock In \emph{ECCV}, 2022.

\bibitem[Karras et~al.(2018)Karras, Aila, Laine, and Lehtinen]{progan}
Tero Karras, Timo Aila, Samuli Laine, and Jaakko Lehtinen.
\newblock Progressive growing of gans for improved quality, stability, and variation.
\newblock \emph{ICLR}, 2018.

\bibitem[Karras et~al.(2019)Karras, Laine, and Aila]{stylegan}
Tero Karras, Samuli Laine, and Timo Aila.
\newblock A style-based generator architecture for generative adversarial networks.
\newblock In \emph{CVPR}, 2019.

\bibitem[Karras et~al.(2020)Karras, Laine, Aittala, Hellsten, Lehtinen, and Aila]{stylegan2}
Tero Karras, Samuli Laine, Miika Aittala, Janne Hellsten, Jaakko Lehtinen, and Timo Aila.
\newblock Analyzing and improving the image quality of stylegan.
\newblock In \emph{CVPR}, 2020.

\bibitem[Kim et~al.(2022)Kim, Kwon, and Ye]{kim2022diffusionclip}
Gwanghyun Kim, Taesung Kwon, and Jong~Chul Ye.
\newblock Diffusionclip: Text-guided diffusion models for robust image manipulation.
\newblock In \emph{CVPR}, 2022.

\bibitem[Kingma and Welling(2013)]{kingma2013auto}
Diederik~P Kingma and Max Welling.
\newblock Auto-encoding variational bayes.
\newblock \emph{ICLR14}, 2013.

\bibitem[Li et~al.(2019)Li, Zhang, and Malik]{IMLE}
Ke Li, Tianhao Zhang, and Jitendra Malik.
\newblock Diverse image synthesis from semantic layouts via conditional imle.
\newblock In \emph{ICCV}, 2019.

\bibitem[Liu et~al.(2022)Liu, Yang, Bi, Xiao, Li, and Gao]{lnp}
Bo Liu, Fan Yang, Xiuli Bi, Bin Xiao, Weisheng Li, and Xinbo Gao.
\newblock Detecting generated images by real images.
\newblock In \emph{ECCV}, 2022.

\bibitem[Marra et~al.(2019)Marra, Gragnaniello, Verdoliva, and Poggi]{marra2019gans}
Francesco Marra, Diego Gragnaniello, Luisa Verdoliva, and Giovanni Poggi.
\newblock Do gans leave artificial fingerprints?
\newblock In \emph{MIPR}, 2019.

\bibitem[Nichol et~al.(2022)Nichol, Dhariwal, Ramesh, Shyam, Mishkin, McGrew, Sutskever, and Chen]{glide}
Alex Nichol, Prafulla Dhariwal, Aditya Ramesh, Pranav Shyam, Pamela Mishkin, Bob McGrew, Ilya Sutskever, and Mark Chen.
\newblock Glide: Towards photorealistic image generation and editing with text-guided diffusion models.
\newblock \emph{ICML}, 2022.

\bibitem[Ojha et~al.(2023)Ojha, Li, and Lee]{ufd}
Utkarsh Ojha, Yuheng Li, and Yong~Jae Lee.
\newblock Towards universal fake image detectors that generalize across generative models.
\newblock In \emph{CVPR}, 2023.

\bibitem[Park et~al.(2019)Park, Liu, Wang, and Zhu]{gaugan}
Taesung Park, Ming-Yu Liu, Ting-Chun Wang, and Jun-Yan Zhu.
\newblock Semantic image synthesis with spatially-adaptive normalization.
\newblock In \emph{CVPR}, 2019.

\bibitem[Radford et~al.(2021)Radford, Kim, Hallacy, Ramesh, Goh, Agarwal, Sastry, Askell, Mishkin, Clark, et~al.]{clip}
Alec Radford, Jong~Wook Kim, Chris Hallacy, Aditya Ramesh, Gabriel Goh, Sandhini Agarwal, Girish Sastry, Amanda Askell, Pamela Mishkin, Jack Clark, et~al.
\newblock Learning transferable visual models from natural language supervision.
\newblock In \emph{ICML}, 2021.

\bibitem[Ramesh et~al.(2021)Ramesh, Pavlov, Goh, Gray, Voss, Radford, Chen, and Sutskever]{dalle}
Aditya Ramesh, Mikhail Pavlov, Gabriel Goh, Scott Gray, Chelsea Voss, Alec Radford, Mark Chen, and Ilya Sutskever.
\newblock Zero-shot text-to-image generation.
\newblock In \emph{ICCV}, 2021.

\bibitem[Rombach et~al.(2022)Rombach, Blattmann, Lorenz, Esser, and Ommer]{ldm}
Robin Rombach, Andreas Blattmann, Dominik Lorenz, Patrick Esser, and Bj{\"o}rn Ommer.
\newblock High-resolution image synthesis with latent diffusion models.
\newblock In \emph{CVPR}, 2022.

\bibitem[Rossler et~al.(2019)Rossler, Cozzolino, Verdoliva, Riess, Thies, and Nie{\ss}ner]{deepfakes}
Andreas Rossler, Davide Cozzolino, Luisa Verdoliva, Christian Riess, Justus Thies, and Matthias Nie{\ss}ner.
\newblock Faceforensics++: Learning to detect manipulated facial images.
\newblock In \emph{ICCV}, 2019.

\bibitem[Ruiz et~al.(2023)Ruiz, Li, Jampani, Pritch, Rubinstein, and Aberman]{ruiz2023dreambooth}
Nataniel Ruiz, Yuanzhen Li, Varun Jampani, Yael Pritch, Michael Rubinstein, and Kfir Aberman.
\newblock Dreambooth: Fine tuning text-to-image diffusion models for subject-driven generation.
\newblock In \emph{CVPR}, 2023.

\bibitem[Saharia et~al.(2022)Saharia, Chan, Saxena, Li, Whang, Denton, Ghasemipour, Gontijo~Lopes, Karagol~Ayan, Salimans, et~al.]{saharia2022photorealistic}
Chitwan Saharia, William Chan, Saurabh Saxena, Lala Li, Jay Whang, Emily~L Denton, Kamyar Ghasemipour, Raphael Gontijo~Lopes, Burcu Karagol~Ayan, Tim Salimans, et~al.
\newblock Photorealistic text-to-image diffusion models with deep language understanding.
\newblock \emph{NeurIPS}, 2022.

\bibitem[Schuhmann et~al.(2021)Schuhmann, Vencu, Beaumont, Kaczmarczyk, Mullis, Katta, Coombes, Jitsev, and Komatsuzaki]{schuhmann2021laion}
Christoph Schuhmann, Richard Vencu, Romain Beaumont, Robert Kaczmarczyk, Clayton Mullis, Aarush Katta, Theo Coombes, Jenia Jitsev, and Aran Komatsuzaki.
\newblock Laion-400m: Open dataset of clip-filtered 400 million image-text pairs.
\newblock \emph{NeurIPS Workshop}, 2021.

\bibitem[Sha et~al.(2023)Sha, Li, Yu, and Zhang]{defake}
Zeyang Sha, Zheng Li, Ning Yu, and Yang Zhang.
\newblock De-fake: Detection and attribution of fake images generated by text-to-image generation models.
\newblock In \emph{CCS}, 2023.

\bibitem[Song et~al.(2021)Song, Sohl-Dickstein, Kingma, Kumar, Ermon, and Poole]{song2020score}
Yang Song, Jascha Sohl-Dickstein, Diederik~P Kingma, Abhishek Kumar, Stefano Ermon, and Ben Poole.
\newblock Score-based generative modeling through stochastic differential equations.
\newblock \emph{ICLR}, 2021.

\bibitem[Tan et~al.(2023)Tan, Zhao, Wei, Gu, and Wei]{lgrad}
Chuangchuang Tan, Yao Zhao, Shikui Wei, Guanghua Gu, and Yunchao Wei.
\newblock Learning on gradients: Generalized artifacts representation for gan-generated images detection.
\newblock In \emph{CVPR}, 2023.

\bibitem[Wang et~al.(2020)Wang, Wang, Zhang, Owens, and Efros]{cnndet}
Sheng-Yu Wang, Oliver Wang, Richard Zhang, Andrew Owens, and Alexei~A Efros.
\newblock Cnn-generated images are surprisingly easy to spot... for now.
\newblock In \emph{CVPR}, 2020.

\bibitem[Wang et~al.(2023)Wang, Bao, Zhou, Wang, Hu, Chen, and Li]{dire}
Zhendong Wang, Jianmin Bao, Wengang Zhou, Weilun Wang, Hezhen Hu, Hong Chen, and Houqiang Li.
\newblock Dire for diffusion-generated image detection.
\newblock In \emph{ICCV}, 2023.

\bibitem[Yang et~al.(2024)Yang, Wang, Qian, Zhu, and Wu]{cow}
Yongqi Yang, Ruoyu Wang, Zhihao Qian, Ye Zhu, and Yu Wu.
\newblock Diffusion in diffusion: Cyclic one-way diffusion for text-vision-conditioned generation.
\newblock \emph{ICLR}, 2024.

\bibitem[Yu et~al.(2015)Yu, Seff, Zhang, Song, Funkhouser, and Xiao]{yu2015lsun}
Fisher Yu, Ari Seff, Yinda Zhang, Shuran Song, Thomas Funkhouser, and Jianxiong Xiao.
\newblock Lsun: Construction of a large-scale image dataset using deep learning with humans in the loop.
\newblock \emph{Arxiv}, 2015.

\bibitem[Zeiler and Fergus(2014)]{occlusion}
Matthew~D. Zeiler and Rob Fergus.
\newblock Visualizing and understanding convolutional networks.
\newblock In \emph{ECCV}, 2014.

\bibitem[Zhang et~al.(2023)Zhang, Rao, and Agrawala]{controlnet}
Lvmin Zhang, Anyi Rao, and Maneesh Agrawala.
\newblock Adding conditional control to text-to-image diffusion models.
\newblock In \emph{ICCV}, 2023.

\bibitem[Zhang et~al.(2019)Zhang, Karaman, and Chang]{autogan}
Xu Zhang, Svebor Karaman, and Shih-Fu Chang.
\newblock Detecting and simulating artifacts in gan fake images.
\newblock In \emph{WIFS}, 2019.

\bibitem[Zhu et~al.(2017)Zhu, Park, Isola, and Efros]{cyclegan}
Jun-Yan Zhu, Taesung Park, Phillip Isola, and Alexei~A Efros.
\newblock Unpaired image-to-image translation using cycle-consistent adversarial networks.
\newblock In \emph{ICCV}, 2017.

\bibitem[Zhu et~al.(2024{\natexlab{a}})Zhu, Chen, Yan, Huang, Lin, Li, Tu, Hu, Hu, and Wang]{genimage}
Mingjian Zhu, Hanting Chen, Qiangyu Yan, Xudong Huang, Guanyu Lin, Wei Li, Zhijun Tu, Hailin Hu, Jie Hu, and Yunhe Wang.
\newblock Genimage: A million-scale benchmark for detecting ai-generated image.
\newblock \emph{NeurIPS}, 2024{\natexlab{a}}.

\bibitem[Zhu et~al.(2024{\natexlab{b}})Zhu, Wu, Deng, Russakovsky, and Yan]{zhu2024boundary}
Ye Zhu, Yu Wu, Zhiwei Deng, Olga Russakovsky, and Yan Yan.
\newblock Boundary guided learning-free semantic control with diffusion models.
\newblock \emph{NeurIPS}, 2024{\natexlab{b}}.

\end{thebibliography}
}
\clearpage
\appendix
\setcounter{page}{1}
\setcounter{section}{0} 
\setcounter{figure}{0}  
\setcounter{table}{0}  
\maketitlesupplementary

In the appendix, we add some experiments to delve deeper into the robustness of the differentiated features. We conduct a sensitivity analysis on the only variable of the patch-shuffle operation, the patch size, in Sec.~\ref{app:granularity}; we use different classification heads to demonstrate the robustness of the differentiated features to the classification heads in Sec.~\ref{app:different_heads}; we showcase the samples that were correctly identified additionally by using differentiated features compared to the baseline UFD in Sec.~\ref{app:additional_detected_samples}; we detailly report the test results of our method and the baseline across various generators in Sec.~\ref{app:detailed_results}; finally, we visualize the focus of UFD and $D^3$ to demonstrate $D^3$'s ability of capturing more comprehensive and universal artifacts in Sec.~\ref{app:attention_visualization}.

\section{Sensitivity of Discrepancy Features to Patch Size}
\label{app:granularity}
\vspace{-0.15in}
\begin{table}[!hp]
    \centering
    \tabcolsep=1pt
    \resizebox{0.45\textwidth}{!}{
    \begin{tabular}{c|cccccc}
    \toprule
        \multirow{3}{*}{Patch size} & \multicolumn{6}{|c}{Validataion} \\
        ~ & \multicolumn{2}{|c}{In-domain} & \multicolumn{2}{c}{Out-of-domain} & \multicolumn{2}{c}{Total} \\  
        ~ & Mean acc. & AP & Mean acc. & AP & Mean acc. & AP \\
    \hline
        1        & 0.958 & 0.992 & 0.837 & 0.934 & 0.885 & 0.960   \\
        14       & 0.967 & 0.995 & 0.859 & 0.944 & 0.904 & 0.968   \\
        28       & 0.966 & 0.995 & 0.871 & 0.939 & 0.909 & 0.965   \\
        56       & 0.962 & 0.998 & 0.871 & 0.943 & 0.907 & 0.965   \\ 
        112      & 0.949 & 0.989 & 0.858 & 0.942 & 0.895 & 0.964   \\ 
        224      & 0.889 & 0.964 & 0.829 & 0.935 & 0.853 & 0.946   \\
    \bottomrule
    \end{tabular}}
    \caption{\textbf{Results of different patch sizes on the validation set.} The ablated patch sizes range from 1 to 224 (the image size). The significant improvement brought by the switch from 224 to 112 shows the effectiveness of introducing discrepancy. Patch sizes 14, 28, and 56 achieve similarly high performance. But patch size 1's performance drops for the over-destruction of local artifacts.}
    \label{granularities}
\end{table}

We study how the patch size affects the learning of discrepancy signals.
A tradeoff exists between increasing the discrepancy between features and mining the universal local artifacts, i.e. the smaller patch size offers more discrepancy but retains fewer local artifacts.
Therefore, given the original image size of 224, we conduct validations with different patch sizes, ranging from 1 to 224, to see the changing trend. 
These experiments adhere to the previous setting, with only patch size being adjusted.
As shown in Tab.~\ref{granularities}, changing the patch size from 224 to 112 brings a significant improvement of 6.0 points in ID accuracy and 2.9 points in OOD accuracy, suggesting that additional discrepancy in features helps in expanding the representation of features and extracting universal artifacts.
The patch sizes 14, 28, and 56 yield similarly high overall performance, showing the introduced discrepancy's robustness in different patch sizes. 
Note that when the patch size is 1, the local artifacts of the shuffled image are significantly affected, resulting in a drop in model performance compared to patch size 14.
In our SOTA version, we directly opt for a patch size of 14 to align with the backbone CLIP:ViT-L/14~\cite{clip} while introducing the highest discrepancy in features.

\section{Different Classifier Heads}
\label{app:different_heads}
\vspace{-0.15in}

\begin{table}[!hp]
    \centering
    \tabcolsep=1pt

    \resizebox{0.45\textwidth}{!}{
    \begin{tabular}{c|cccccc}
    \toprule
        \multirow{3}{*}{Architecture} & \multicolumn{6}{|c}{Validataion} \\
        ~ & \multicolumn{2}{|c}{In-domain} & \multicolumn{2}{c}{Out-of-domain} & \multicolumn{2}{c}{Total} \\  
        ~ & Mean acc. & AP & Mean acc. & AP & Mean acc. & AP \\
    \hline
        FC              & 0.918 & 0.977 & 0.835 & 0.938 & 0.868 & 0.954   \\
        MLP             & 0.960 & 0.995 & 0.865 & 0.932 & 0.903 & 0.963   \\
        Self-Attention  & 0.967 & 0.995 & 0.859 & 0.944 & 0.904 & 0.968   \\
        Transformer     & 0.965 & 0.995 & 0.872 & 0.952 & 0.909 & 0.973   \\ 
    \bottomrule
    \end{tabular}}
    \caption{\textbf{Results of different classifier heads on the validation set.} We evaluate four classifier network architectures and find that a network that learns the correlations between features will perform better.}
    \vspace{-0.1in}
    \label{classifier}
\end{table}

We investigate how different classifier heads influence the model's performance to verify the effectiveness of artifact invariance learning.
We evaluate four architectures: (i) \textbf{FC}: a single fully connected layer, (ii) \textbf{MLP}: A two-layer non-linear perceptron network with ReLU activation and a hidden layer dimension of $2 \times 1024$ neurons, (iii) \textbf{Self-Attention}: a network consisting of a self-attention layer \cite{vit} and a single fully connected layer, and (iv) \textbf{Transformer}: A network composed of two transformer encoder layers with 4 attention heads and a forward dimension of $4 \times 1024$ \cite{vit} and one fully connected layer. 

Tab.~\ref{classifier} presents the results of these variants in our proposed experimental setting. 
The findings show that the results of MLP, Self-Attention, and Transformer are significantly improved compared to FC. This means establishing the correlations between the two discrepancy features helps learn universal artifacts.
In addition, the performances of Self-Attention, MLP, and Transformer don't show an obvious gap,  which demonstrates that our discrepancy features are highly distinguishable for deepfake detection.

\begin{figure*}[!tp]
    \centering
    \setlength{\abovecaptionskip}{3pt}
    \includegraphics[width=0.98\textwidth]{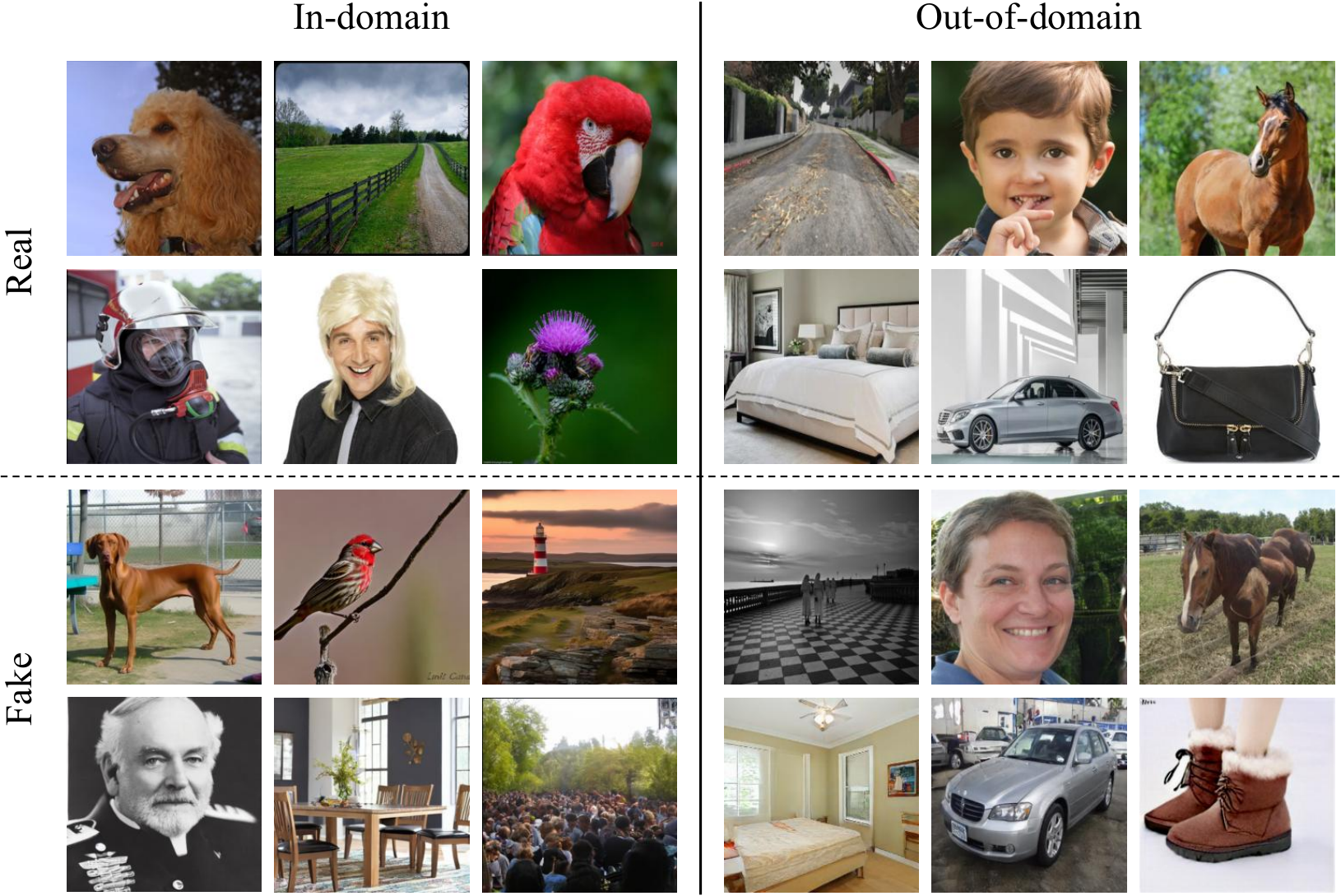}
    \caption{\textbf{Visualization of uniquely detected samples.} We present a selection of samples from both in-domain and out-of-domain, that are accurately classified by our approach, yet erroneously classified by the UFD~\cite{ufd}, with a discrepancy in classification confidence exceeding 0.8.}
    \label{apx:visualization}
\end{figure*}

\begin{table*}[!h]

    \resizebox{0.99\linewidth}{!}{
    \centering
    \begin{tabular}{cccccccccccccccccccccccc}
        \toprule
        ~ & \multicolumn{8}{c}{In-domain} & \multicolumn{12}{c}{Out-of-domain} & ~ & ~ & ~  \\
        \cmidrule(lr){2-9}\cmidrule(lr){10-21}
         ~ & ADM  & Big & GLI & Mid & LDM & VQ & wu & Pro & Cycle & Style & Style & Gau & Star & Deep & which- & SI & SAN & CRN & IM & DAL & ~ & ~ & ~ \\
        methods & ~ & GAN & DE & jour & ~  & DM & kong & GAN & GAN & GAN & GAN2 & GAN & GAN & fakes & face & TD & ~ & ~ & LE & L$\cdot$E & ID & OOD & Total \\
        ~ & ~ & ~ & ~ & ney & ~ & ~ & ~ & ~ & ~ & ~ & ~ & ~ & ~ & ~ & isreal & ~ & ~ & ~ & ~ & ~ & ~ & ~ & ~ \\
        \midrule
        CNNDet & 89.6 & 79.1 & 98.2 & 97.3 & 93.7 & 97.0 & 95.7 & 95.6 & 72.0 & 75.7 & 80.0 & 56.4 & 98.2 & 80.2 & 47.3 & 49.7 & 72.4 & 56.0 & 56.0 & 94.7 & 93.3 & 69.9 & 79.2  \\
        Patchfor & 99.8 & 85.1 & 99.6 & 99.6 & 99.7 & 99.7 & 99.7 & 99.9 & 93.2 & 98.1 & 94.4 & 59.1 & 99.8 & 90.7 & 61.6 & 65.4 & 84.5 & 50.0 & 50.0 & 99.6 & 97.9 & 78.9 & 86.5  \\
        LNP & 91.0 & 72.5 & 95.0 & 94.4 & 92.2 & 89.4 & 92.8 & 87.3 & 71.0 & 89.9 & 85.0 & 68.8 & 83.7 & 65.5 & 53.9 & 52.0 & 55.8 & 50.1 & 62.6  & 84.6 & 89.3 & 68.6 & 76.9 \\
        DIRE & 99.9 & 82.6 & 99.9 & 99.8 & 99.9 & 100 & 100 & 98.4 & 60.3 & 94.2 & 95.1 & 55.3 & 88.6 & 67.9 & 50.0 & 50.0 & 59.7 & 50.0 & 50.0 & 99.7 & 97.6 & 68.4 & 80.1  \\
        UFD & 83.2 & 92.0 & 86.3 & 80.2 & 85.4 & 89.0 & 85.6 & 91.1 & 74.8 & 79.4 & 82.4 & 95.5 & 89.0 & 75.9 & 79.8 & 73.1 & 61.9 & 87.9 & 90.0  & 86.6 & 86.6 & 81.4 & 83.5  \\
        Ours & 94.8 & 98.5 & 95.0 & 96.8 & 94.4 & 96.7 & 97.1 & 99.4 & 92.7 & 94.9 & 95.7 & 98.1 & 96.0 & 67.7 & 83.1 & 73.8 & 62.6 & 88.1 & 95.0 & 92.8 & 96.6 & 87.6 & 90.7  \\
        \bottomrule
    \end{tabular}
    }
    \caption{\textbf{Detailed mean accuracy results of comparisons with the state-of-the-art on the testing set.} We report the mean accuracy per generator in the percentage form. The results of generators with the same architecture but with different parameters are averaged.}
    \label{apx:detailedcomparisons}
\end{table*}
\section{Comparative Analysis of Uniquely Detected Samples}

\label{app:additional_detected_samples}
We take a further step to explore how our method outperforms. We present a selection of samples from both in-domain and out-of-domain. These samples are accurately classified by our approach, yet erroneously classified by the UFD~\cite{ufd}, with a discrepancy in classification confidence exceeding 0.8. As shown in Fig.\ref{apx:visualization}, these samples are challenging to discern with the naked eye. This compellingly demonstrates that our method is capable of learning deeper and more universal artifacts, thereby retaining its effectiveness even when confronted with such challenging samples.

\section{Detailed Mean Accuracy Results of Comparisons with the State-of-the-arts}
\label{app:detailed_results}
\label{apx:comparisons}
In this section, we report the detailed mean accuracy results of comparisons with the state-of-the-art in Sec.~\ref{apx:detailedcomparisons}, as a supplement to Table 1 in Sec 4.5, including ADM~\cite{guideddiffusion}, BigGAN~\cite{biggan}, GLIDE~\cite{glide}, Midjourney~\cite{midjourney}, LDM~\cite{ldm}, VQDM~\cite{vqdm}, wukong~\cite{wukong}, ProGAN~\cite{progan}, CycleGAN~\cite{cyclegan}, StyleGAN~\cite{stylegan}, StyleGAN2~\cite{stylegan2}, GauGAN~\cite{gaugan}, StarGAN~\cite{stargan}, Deepfakes~\cite{deepfakes}, whichfaceisreal~\cite{whichface}, SITD~\cite{SITD}, SAN~\cite{SAN}, CRN~\cite{crn}, IMLE~\cite{IMLE}, and DALL$\cdot$E~\cite{dalle}. Results of generators with the same architecture but different parameters are averaged. For example, the result of BigGAN~\cite{biggan} in this table is the average of BigGAN in UFD~\cite{ufd} and BigGAN in GenImage~\cite{genimage}.

\begin{figure*}[!t]
    \centering
    \setlength{\abovecaptionskip}{3pt}
    \includegraphics[width=\textwidth]{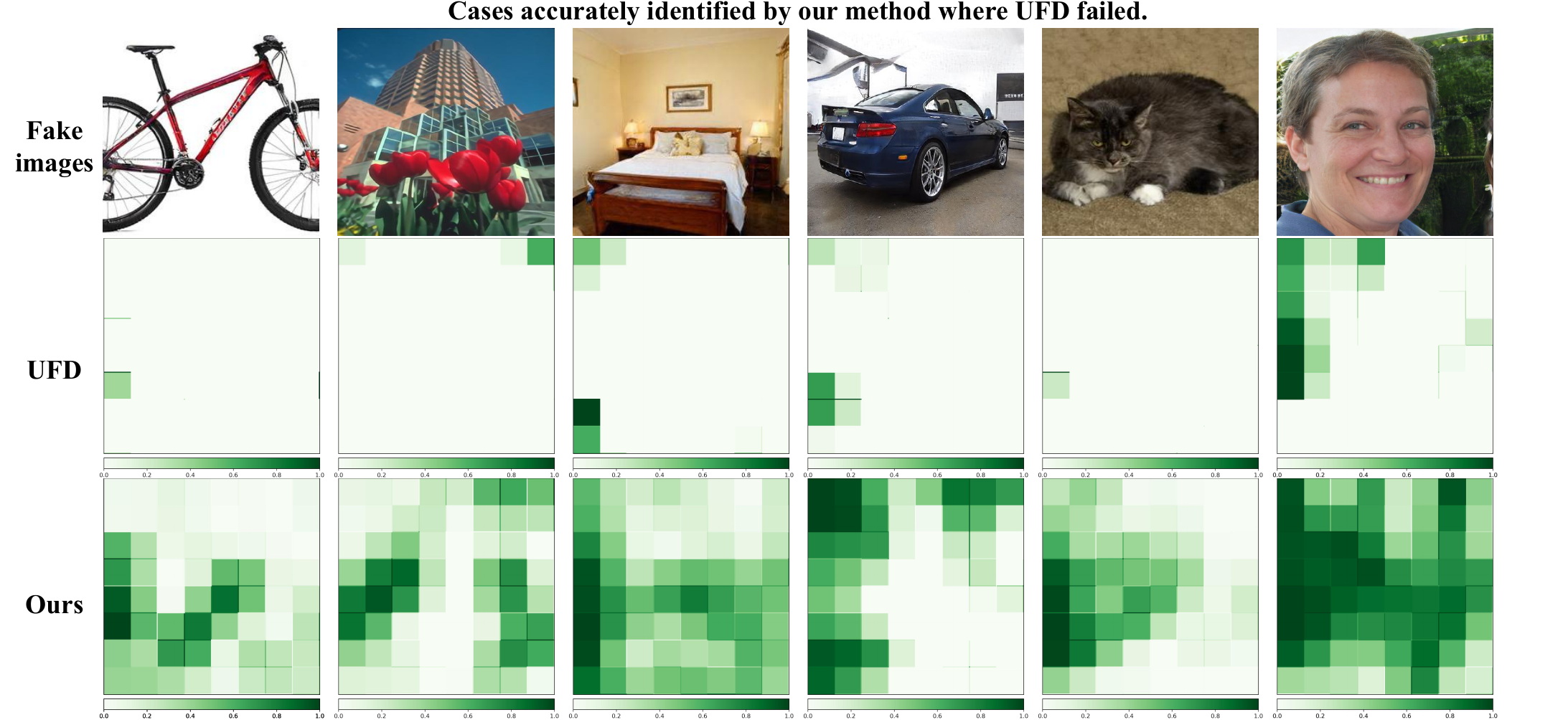}
    \caption{Visualizations of fake image samples along with the corresponding occlusion maps~\cite{occlusion} for each detector. These images represent cases accurately identified by $D^3$ where UFD failed.}
    \label{fig:atten_visualization}
\end{figure*}

\section{Visualization of the Detector's Attention}
\label{app:attention_visualization}

$D^3$ effectively learns more universal artifacts through discrepancies introduced by image transformations, achieving superior performance. We employ the occlusion technique \cite{occlusion} to visually demonstrate this to identify regions of interest during inference, as illustrated in \cref{fig:atten_visualization}. Our analysis reveals that UFD focuses on limited regions, often leading to misjudgments due to its narrow attention. In contrast, $D^3$ exhibits a significantly broader attention range, further validating that the introduced discrepancy enables D3 to capture more comprehensive and universal artifacts.


\end{document}